\documentclass{article}
\newif\ifarxiv
\arxivtrue  
\usepackage[final,main]{neurips_2026}
\usepackage[utf8]{inputenc}
\usepackage[T1]{fontenc}
\usepackage{courier}
\usepackage{graphicx}
\usepackage{hyperref}
\usepackage{url}
\usepackage{booktabs}
\usepackage{amsmath,amsfonts,amssymb}
\usepackage{nicefrac}
\usepackage{microtype}
\usepackage[dvipsnames]{xcolor}
\usepackage{enumitem}
\usepackage{placeins}
\usepackage{array,multirow}
\title{MSP: Modality Self-Play for Training Multimodal LLMs Without Paired Cross-Modal Data}
\author{
  Robert Moseley \\
  University of California, \\
  Riverside \\
  {\small\texttt{rmose009@ucr.edu}}
  \And
  Md Kaykobad Reza \\
  University of California, \\
  Riverside \\
  {\small\texttt{mreza025@ucr.edu}}
  \And
  Ameya Patil \\
  Amazon \\
  {\small\texttt{ameyapat@amazon.com}}
  \AND
  Edward Ayrapetian \\
  Amazon \\
  {\small\texttt{edayrap@amazon.com}}
  \And
  Salman Asif \\
  University of California, \\
  Riverside \\
  {\small\texttt{sasif@ucr.edu}}
}

\hypersetup{
  hidelinks,
  pdftitle={MSP: Modality Self-Play for Training Multimodal LLMs Without Paired Cross-Modal Data},
  pdfauthor={Robert Moseley; Md Kaykobad Reza; Ameya Patil; Edward Ayrapetian; Salman Asif},
  pdfsubject={NeurIPS 2026}
}
\begin{document}

\maketitle

\begin{abstract}
Large multimodal models have achieved remarkable progress by scaling model capacity and training on massive paired multimodal data. However, this paradigm introduces a fundamental bottleneck: aligned multimodal datasets are expensive to curate, difficult to scale across diverse modality combinations, and increasingly constrain further progress. We introduce \emph{Modality Self-Play} (MSP), a training framework for multimodal foundation models based on proxy-mediated modality composition. Instead of relying on paired cross-modal data, MSP uses anchor-supporting proxy textual descriptions of unseen modalities as semantic bridges during training. Given an observed modality (e.g., image or audio), the model generates proxy descriptions for complementary unseen modalities, which are then encoded and aligned with the observed modality through learned bridges, enabling cross-modal interaction without real paired non-text supervision. 
Our central hypothesis is that proxy text, combined with contrastively trained pretrained encoders, provides sufficient semantic structure for multimodal composition to emerge. We show that MSP enables zero-shot modality composition: models trained only with a real anchor modality and accompanying proxy descriptions can integrate and reason over multiple modalities at inference time despite never observing real paired non-text data during training. Empirically, MSP achieves state-of-the-art or competitive performance on multiple multimodal benchmarks. Ablation studies further demonstrate that learned bridges improve alignment, while anchor-supporting proxy descriptions enable effective cross-modal composition. MSP removes the need for real paired non-text (e.g., audio-vision) supervision, while continuing to rely on modality-text pairs, pretrained contrastive encoders, and modality-distribution bridge calibration.
\end{abstract}

\section{Introduction}
\label{sec:introduction}

Large multimodal models (LMMs) \citep{huang2024multimodalSurvey, wu2023multimodalSurvey} have advanced the frontier of intelligent systems by enabling unified reasoning over language, vision, audio, and other heterogeneous modalities. Much of this progress has been driven by scaling: larger models, broader modality coverage, and increasingly large paired multimodal datasets have led to strong gains across perception and reasoning tasks \citep{qwen2025omni, baichuan2025omni15}. However, most existing approaches rely on a common assumption: effective cross-modal reasoning requires explicit supervision from paired multimodal data.
This dependence on paired multimodal data presents a key scalability challenge. High-quality aligned multimodal datasets are expensive to collect, difficult to scale across modality combinations, and often domain-specific. As models expand to incorporate more modalities, the cost of acquiring comprehensive paired data grows combinatorially. As a result, progress in multimodal learning is increasingly constrained by the availability of paired multimodal datasets rather than model design.

In contrast, large amounts of bimodal data (e.g., vision-language or audio-language) are readily available on open source repositories. These datasets contain rich semantic information, raising an important question: \emph{can models learn to reason over multiple modalities without ever observing real paired non-text data during training?} Prior work has explored related directions through shared encoders \citep{han2024onellm}, modular model design \citep{su2023pandagpt, han2023imagebindllm}, and weak supervision \citep{panagopoulou2023x-instructblip}. However, most approaches still rely on some form of paired multimodal supervision during training \citep{wu2024nextgpt} or exhibit limited performance on standard multimodal benchmarks \citep{zhao2023chatbridge, han2024onellm, panagopoulou2023x-instructblip}.

In this work, we explore an alternative approach based on \emph{proxy-mediated modality alignment}. We introduce \textbf{Modality Self-Play (MSP)}, a training paradigm that replaces real paired non-text supervision with proxy textual descriptions. Given an observed modality (e.g., an image), MSP generates proxy text describing complementary unseen modalities (e.g., audio). These proxy descriptions are encoded using contrastively pretrained text encoders (e.g., CLIP/CLAP-style encoders \citep{radford2021clip, wu2023clap}) and projected to the shared language embedding space via a bridge and MLP projector. During training, the model never observes real paired non-text data; instead, it learns from single modalities paired with proxy descriptions that act as semantic bridges. This design leverages the fact that contrastively pretrained encoders provide transferable cross-modal structure \citep{girdhar2023imagebind}. Proxy text does not replicate unseen modalities, but provides a shared semantic space through which representations can be aligned. We therefore do not assume the absence of cross-modal priors, but rather eliminate the need for real paired non-text data during training.

Our central hypothesis is that proxy-mediated alignment is sufficient for multimodal composition to emerge, enabling LMMs to reason jointly over multiple modalities without real paired non-text data during training.
We evaluate this hypothesis empirically across multiple multimodal reasoning benchmarks. We show that MSP enables \emph{zero-shot modality composition}: models trained using single modalities and proxy descriptions can reason over multiple modalities at inference time despite never observing them jointly during training. MSP achieves \emph{competitive or state-of-the-art performance} on standard multimodal benchmarks \citep{yang2022avqa, li2022musicavqa, li2024omnibench} without using real paired non-text training data (Section~\ref{sec:exp_headline}). We further analyze the role of proxy descriptions and alignment mechanisms through ablation studies, showing that learned bridges improve alignment between proxy embeddings and the corresponding modality representations, resulting in improved multimodal reasoning performance (Section~\ref{sec:proxy_contribution}).
\textbf{Our contributions are summarized as follows:}
\begin{itemize}[leftmargin=*]
    \item We introduce \textbf{Modality Self-Play (MSP)}, a training framework that replaces real paired non-text supervision with proxy-mediated alignment.
    \item We show that proxy textual descriptions can serve as semantic bridges, enabling zero-shot multimodal composition without real paired non-text training data. MSP achieves competitive or state-of-the-art performance on multiple multimodal benchmarks (Section~\ref{sec:exp_headline}).
    \item Ablation studies show that bridges improve alignment between proxy and modality representations, leading to stronger multimodal reasoning performance without real paired non-text data (Section~\ref{sec:proxy_contribution}).
\end{itemize}

MSP removes the need for real paired non-text (e.g., audio-vision) supervision, while continuing to rely on modality-text pairs, pretrained contrastive encoders, and modality-distribution bridge calibration.

\section{Related Work}
\label{sec:related-work}

Our work relates to three lines of research in multimodal foundation models: (i) multimodal LLMs (MLLMs) trained with paired multimodal data, (ii) approaches that align independently observed modalities to a shared language embedding space, and (iii) text-mediated proxy methods that use language as an intermediate representation.

\textbf{Training MLLMs with paired multimodal data.}
The dominant paradigm for MLLMs relies on training with explicitly paired multimodal data, where language or instruction prompts are aligned with one or more observed modalities. Early vision-language models such as BLIP-2 \citep{li2023blip2} and LLaVA \citep{liu2023llava} demonstrate strong performance through large-scale image-text pretraining and instruction fine-tuning. This paradigm has been extended to broader modality coverage: NExT-GPT \citep{wu2024nextgpt}, AnyGPT \citep{zhan2024anygpt}, Unified-IO~2 \citep{lu2024unified}, and MIO \citep{wang2024mio} integrate multiple modalities within unified architectures, while recent omni-modal systems such as Qwen2.5-Omni \citep{qwen2025omni}, Baichuan-Omni-1.5 \citep{baichuan2025omni15}, and MiniCPM \citep{yao2024minicpm} scale this approach with large multimodal training pipelines.
While effective, these methods depend on the availability of aligned paired multimodal datasets, which can be costly to curate and difficult to scale across modality combinations. In contrast, our work explores whether multimodal composition can be learned without requiring explicit paired multimodal supervision.

\textbf{Multimodal learning via modality-text alignment.}
A growing line of work reduces reliance on direct multimodal pairing by aligning each modality to language or to a shared embedding space. ImageBind-LLM \citep{han2023imagebindllm} leverages pretrained modality-specific encoders to align multiple modalities to a shared image embedding space before connecting them to a large language model. PandaGPT \citep{su2023pandagpt}, X-InstructBLIP \citep{panagopoulou2023x-instructblip}, and OneLLM \citep{han2024onellm} extend this approach by aligning multiple modalities to a shared language embedding space through projectors and modality-specific adapters. ChatBridge \citep{zhao2023chatbridge} utilizes language as an intermediate representation connecting modalities, reducing the need for direct modality-to-modality pairing.
These approaches demonstrate that language can serve as a unifying interface for multimodal learning. However, they often struggle with cross-modal composition at inference time, since the models are not explicitly trained to reason jointly over multiple modalities. In contrast, MSP introduces proxy-mediated alignment that incorporates compositional signals during training without requiring real paired non-text datasets.

\textbf{Text-mediated proxies and representation bridging.}
Several methods use text representations as proxies for other modalities. DeCap \citep{li2023decap} and CapDec \citep{nukrai2022capdec} train decoders using only text embeddings and map visual representations into the text embedding space during inference, enabling zero-shot captioning. Similar ideas have been explored for image, video, and audio captioning \citep{wang2023knight, deshmukh2024noaudio, li2025drcap}. These approaches demonstrate that text can act as an intermediate representation when multimodal data is limited.

However, these methods primarily focus on single-modality generation tasks and do not investigate whether text proxies can be used to learn multimodal composition without paired multimodal data. MSP differs from existing approaches in two important ways. First, proxy text conditioned on an observed modality is generated and used during training to guide cross-modal alignment, rather than serving as a substitute for unseen modalities at inference time. Second, these proxies are used to train the model for cross-modal reasoning by using them as semantic surrogates for unseen modalities within the LLM. This positions MSP as a training-time framework for inducing multimodal composition through proxy-mediated alignment.

\section{Method}
\label{sec:method}

We introduce \emph{Modality Self-Play} (MSP), a training framework that enables MLLMs to reason jointly over audio and vision at inference time without observing any real audio-vision pairs during training. We define \emph{paired cross-modal data} as samples containing two or more non-text modalities originating from the same underlying source (e.g., audio and vision from the same recording). MSP uses standard modality-text supervision (e.g., image-language or audio-language) to train modality-specific projectors and calibrate the alignment bridges described below, but never uses paired non-text modality data. We first describe the model architecture in Section~\ref{sec:model-architecture}, followed by the two training stages: per-modality projector training (Section~\ref{sec:phase1}) and self-play composition training (Section~\ref{sec:phase2}).

\begin{figure}[t]
  \centering
  \includegraphics[width=\linewidth]{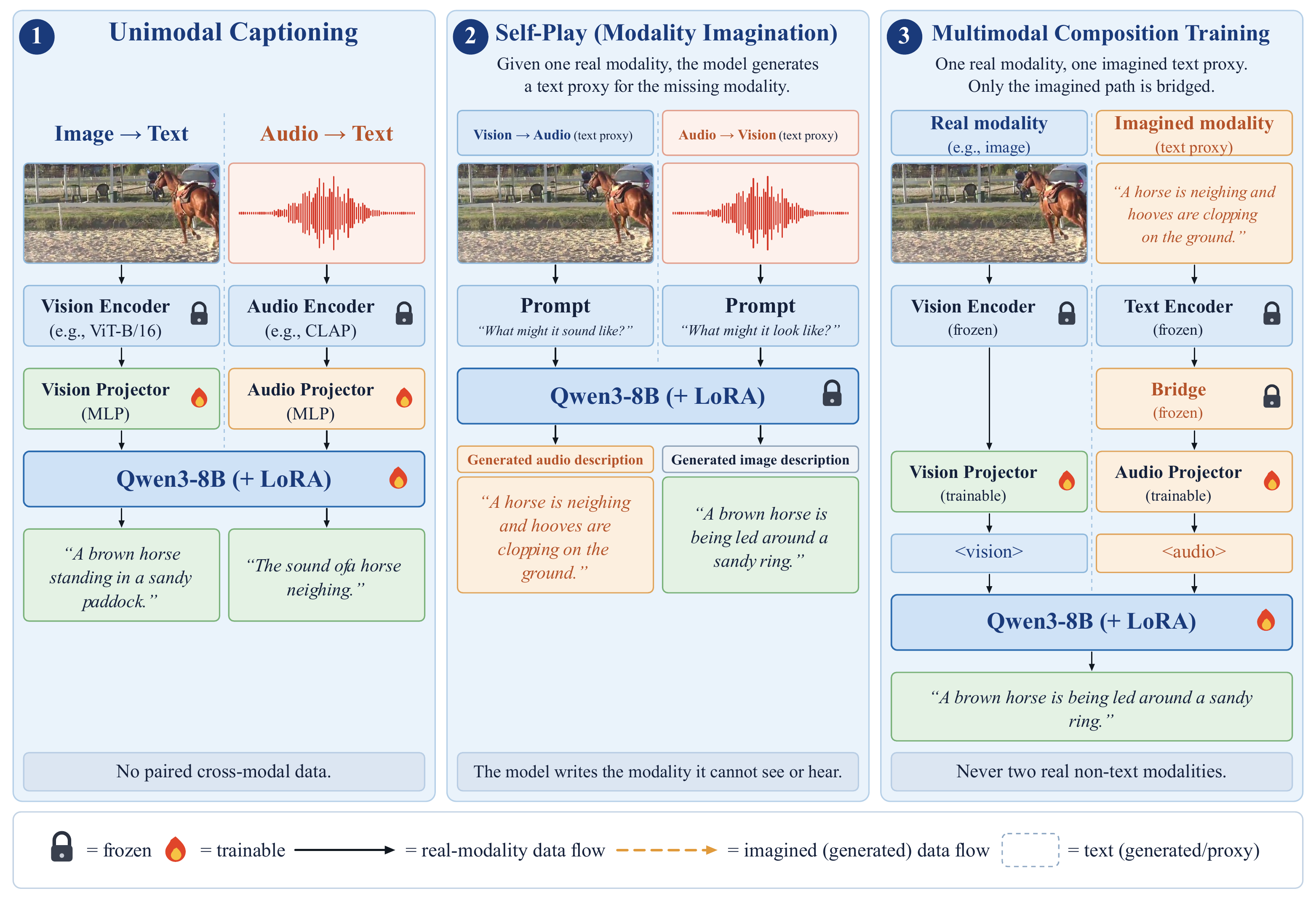}
  \caption{{MSP training pipeline.} \textbf{(1)}~Phase~1 trains one
  projector per modality using modality--text pairs only.
  \textbf{(2)}~The frozen Phase-1 model then generates a text proxy for
  the modality it does not observe. \textbf{(3)}~Phase~2 trains on one
  real modality together with the proxy for the missing one. No training
  sample ever contains two real non-text modalities. A component-level
  view of the same architecture, with the bridge formulation and the full
  token sequence, is given in Figure~\ref{fig:msp_architecture_detail}.}
  \label{fig:msp_architecture}
\end{figure}

\subsection{Model Architecture}
\label{sec:model-architecture}

Figure~\ref{fig:msp_architecture} summarizes the full training pipeline, and Figure~\ref{fig:msp_architecture_detail} illustrates the four main components of MSP: modality-specific encoders, modality-specific projectors, text-to-modality bridges, and the language model.

\textbf{Modality-specific encoders.}
We use frozen contrastive encoders to map each non-text modality into a fixed-dimensional feature representation. Specifically, we use LAION-CLAP \citep{wu2023clap} for audio and MetaCLIP ViT-B/16 \citep{xu2024metaclip} for vision. Each encoder produces a single $512$-dimensional feature embedding per input. The encoders remain frozen throughout training to ensure that both real and proxy representations reside in a consistent embedding space, enabling the bridge modules to learn meaningful transformations between them.

\textbf{Modality-specific projectors.}
Each modality uses a lightweight trainable MLP projector $P_m: \mathbb{R}^{512} \rightarrow \mathbb{R}^{D}$ to map modality-specific features into the language embedding space, where $D{=}4096$ is the hidden dimension of Qwen3-8B. Given an input $x_m$ from modality $m$, the encoder produces a feature representation $z_m$, which is projected into the language embedding space as
\begin{equation}
  h_m = P_m(E_m(x_m)).
  \label{eq:projection}
\end{equation}
Each modality has its own projector. We represent every modality using a single token, ensuring that both real and proxy features share the same representation format.

\textbf{LLM backbone.}
We use Qwen3-8B \citep{qwen2025qwen3} as the language backbone. The input prompt contains modality-specific special tokens, \texttt{<audio>} and \texttt{<vision>}, corresponding to audio and vision inputs, respectively. During the forward pass, these tokens are replaced with the projected modality embeddings $h_m$, following the standard design used in multimodal LLMs \citep{liu2023llava, li2023blip2}. We apply LoRA adapters to the attention and MLP layers of Qwen3-8B while keeping the remaining LLM parameters frozen.

\subsection{Phase~1: Per-Modality Projector Training}
\label{sec:phase1}

In Phase~1, we train the audio and vision projectors independently using standard modality-caption datasets. For each batch from modality $m \in \{a, v\}$, the input prompt contains a single special token (e.g., \texttt{<audio>} or \texttt{<vision>}) along with the text instruction. 
The corresponding modality-specific feature is computed and projected into the LLM embedding space using \eqref{eq:projection}, which replaces the special token in the prompt. The encoder $E_m$ is frozen and used only for feature extraction, while the projector $P_m$ and the LoRA adapters in the LLM backbone are trained.
Training is performed using standard next-token prediction over the response tokens:
\begin{equation}
  \mathcal{L}_{\mathrm{P1}}
  = - \sum_{j \in \mathcal{R}} \log p_\theta(y_j \mid y_{<j}, q, h_m),
  \label{eq:phase1_loss}
\end{equation}
where $q$ denotes the text prompt, $\mathcal{R}$ denotes response tokens, and both prompt and placeholder tokens are excluded from the loss computation. 
The resulting Phase~1 model serves two roles in Phase~2: it initializes the modality-specific projectors and LoRA adapters, and it is used as the generator for self-play proxy text describing unseen modalities.

\subsection{Phase~2: Self-Play Composition Training}
\label{sec:phase2}

Phase~2 enables the model to learn cross-modal composition between audio and vision without using any real paired non-text data. Since no audio-vision pairs are available, each training example uses a single real modality as an \emph{anchor} (e.g., vision) and a \emph{proxy text description} for the unseen modality (e.g., audio).
The proxy text is first encoded using a contrastive text encoder and then transformed through a learned text-to-modality bridge before being mapped into the LLM embedding space via the modality-specific projector. This allows proxy representations to be injected into the model through the same interface as real modality features.
When preparing our dataset, any data that could be considered ``paired'' by definition (i.e., containing both vision and audio) was exclusively utilized for one modality for a single training sample. If, for example, there was some sample containing both vision and audio, we designated the sample for either being used with its audio and not with its vision, or vice versa.
This means that any given sample would only ever be used with one of its modalities---not one pass with vision, and a later pass with its audio. We refer to this as the one-real-anchor rule.

In the following, we describe the proxy generation process, the text-to-modality bridge that aligns proxy embeddings with real modality representations, and the Phase~2 training objective.

\textbf{Self-Play Proxy Generation.}
For each Phase-2 example, we select one modality as the anchor (e.g., audio) and a different modality as the target (e.g., vision). The Phase-1 model is conditioned on the anchor via its corresponding pathway and prompted to generate a short text description $t'_{m'}$ of the unseen modality $m'$.
We use prompt templates aligned with the original training style of the contrastive encoders (AudioCaps-style for CLAP, COCO-style for MetaCLIP); full prompts are provided in Appendix~\ref{app:proxy_prompts}. Since the proxy is generated from the anchor, it is anchor-supporting by construction. All proxies are generated once before Phase~2 and cached, so no online generation is required during training.

\textbf{Text-to-Modality Bridge.}
\label{sec:bridge-calibration}
The text-encoder and modality-encoder outputs of the contrastive encoders lie in related but shifted regions of a shared embedding space \citep{liang2022mindthegap}. Directly feeding text-encoder features into the projector $P_m$ would lead to a distribution mismatch, since $P_m$ is trained on modality-encoder features during Phase~1.
To address this, we learn a linear bridge $B_m$ that maps the embeddings from the text-encoder into the modality-encoder embedding space. The bridge is trained via ridge regression on 1,000 paired (caption, modality) samples (AudioCaps \citep{audiocaps} for audio and LLaVA-665K \citep{liu2024llava15} for vision), split into 500 training and 500 held-out samples:
\begin{equation}
  W_m = (T^{\top} T + \lambda I)^{-1} T^{\top} R,
  \quad B_m(t) = W_m^{\top} t, \quad \lambda = 1.0,
  \label{eq:ridge_bridge}
\end{equation}
where $T$ and $R$ denote stacked text-tower and modality-tower embeddings, respectively.
The bridge is trained once before Phase~2 and kept fixed thereafter. The resulting bridged proxy feature
$\hat{z}_{m'} = B_{m'}(E^{\mathrm{text}}_{m'}(t'_{m'}))$
is then passed through the projector $P_{m'}$ and inserted at the corresponding modality special token using the same mechanism as real modality inputs. Bridge calibration is evaluated in Appendix~\ref{app:bridge_variants}, and remains frozen during Phase~2 training.

\textbf{Phase~2 Forward Pass.}
Let $m$ be the real anchor modality and $m'$ be the missing target
modality. Phase~2 constructs one real feature,
$h_m=P_m(E_m(x_m))$, and one bridged proxy feature,
$h'_{m'}=P_{m'}(B_{m'}(E^{\mathrm{text}}_{m'}(t'_{m'})))$.
Both vectors have the same LLM hidden width $D$ and are inserted
into the prompt at their reserved modality placeholders. An
audio-anchored example therefore contains real audio at
\texttt{<audio>} and a bridged vision proxy at \texttt{<vision>};
a vision-anchored example reverses the roles. This construction gives
real and proxy inputs identical token shape and insertion semantics.

\textbf{Phase~2 Training and Data.}
Audio anchors are drawn from Clotho-AQA \citep{lipping2022clotho}, AudioCaps \citep{audiocaps}, WavCaps \citep{mei2024wavcaps}, and from the audio tracks of MSRVTT-QA and MSVD-QA training videos \citep{xu2017videoqa}. Vision anchors are drawn from VQAv2 \citep{goyal2017vqav2}, LLaVA-665K \citep{liu2024llava15}, and additional frames from MSRVTT-QA and MSVD-QA. The trainable parameters are the modality-specific projectors and shared LoRA adapters, while the modality-specific encoders and bridges remain frozen. The objective is the same as \eqref{eq:phase1_loss}, now conditioned on both the real anchor and the proxy feature. This enables cross-modal composition during training without requiring any real paired audio--vision data.

\textbf{Inference.}
At inference time, real modalities are encoded and projected through
the same encoder-projector pathway used during training. 
Tables~\ref{tab:logit_lens} and~\ref{tab:linear_probe} demonstrate that despite the LLM's own ability to easily distinguish between proxies and modalities in residual space, the model converges to the same tokens in the output layer indicating that the model is able to use proxies as substitutes for real modalities during training. We do not generate proxies at inference.

Figure~\ref{fig:qualitative_typeA} (Appendix~\ref{app:qualitative_appendix}) shows representative proxy
examples generated for AVQA and MUSIC-AVQA samples. These examples
illustrate the intended role of proxy text: it should preserve the
semantic content of the anchor while occupying the representation
space of the missing modality.

\section{Experiments and Results}
\label{sec:experiments}

\subsection{Dataset and Implementation Details}
\label{sec:exp_setup}

\textbf{Phase~1 data.}
In Phase~1, we train the modality-specific projectors together with the LLM decoder using standard modality-text pairs to align each modality into a shared language embedding space that can be processed by the LLM. We represent each modality using a single token. We use the audio-text pairs from AudioCaps \citep{audiocaps} and WavCaps \citep{mei2024wavcaps} datasets, while image-text pairs are obtained from the LLaVA-665K dataset \citep{liu2024llava15}.

\textbf{Proxy generation and Phase~2 data.}
Self-play proxies are generated after Phase~1 and cached for the remainder of training. For each training sample, the Phase~1 model is conditioned on the observed anchor modality and prompted to generate textual descriptions of the unseen target modality in its native captioning style. These proxy descriptions are then used as semantic surrogates for the missing modalities during subsequent MSP training.
To generate vision proxies, we use audio samples from the Clotho-AQA \citep{lipping2022clotho}, AudioCaps \citep{audiocaps}, and WavCaps \citep{mei2024wavcaps} datasets, together with the audio tracks of MSRVTT-QA and MSVD-QA \citep{xu2017videoqa}. Conversely, to generate audio proxies, we use images from VQAv2 \citep{goyal2017vqav2} and LLaVA-665K \citep{liu2024llava15}, together with frames from MSRVTT-QA and MSVD-QA. For video samples, we choose either audio or uniformly sample 4 frames, encode each with MetaCLIP ViT-B/16, and mean-pool the resulting features into a single 512-dimensional vision token. 

Importantly, MSP never observes paired audio-vision-text modalities during composition training. During Phase~2 training, we alternate between \emph{audio + proxy vision description} and \emph{vision + proxy audio description} samples to encourage proxy-mediated multimodal composition.

\textbf{Bridge calibration.}
Each bridge is a linear map that re-projects text-tower vectors into
the modality-tower space, correcting the offset between a contrastive
encoder's two towers \citep{liang2022mindthegap}. We fit one per
modality using the ridge regression and calibration split introduced in
Section \ref{sec:bridge-calibration} ($\lambda{=}1.0$).
Both bridges are fit once before Phase~2 begins and then frozen.

\textbf{Evaluation benchmarks.}
We evaluate our method on three cross-modal question answering (QA) benchmarks that require joint reasoning over audio and vision to answer each question: \textbf{AVQA}~\citep{yang2022avqa}, \textbf{MUSIC-AVQA}~\citep{li2022musicavqa}, and \textbf{OmniBench}~\citep{li2024omnibench}. AVQA and MUSIC-AVQA focus on audio-visual question answering, while OmniBench extends this setting to audio-visual-text reasoning, enabling evaluation in a broader multimodal context. Here, ``zero-shot'' means we do not finetune specifically for performance on these benchmarks (meaning we do not use the training data for these benchmarks to enhance performance).

\textbf{Implementation details.}
We use Qwen3-8B \citep{qwen2025qwen3} with
LoRA adapters applied to attention and MLP modules with ranks
$r{\in}\{128,256,512\}$. The modality-specific
projectors are MLPs and are fully trainable. Frozen contrastive
encoders: MetaCLIP B/16 \citep{xu2024metaclip} for vision and
LAION-CLAP \citep{wu2023clap} for audio; per-modality features are
pooled to a single $512$-dim token before projection.
Full optimization details, including the projector-warmup schedule and per-rank effective batch sizes, are listed in Appendix~\ref{app:phase1_config}.

\subsection{Performance on Multimodal Benchmarks} 
\label{sec:exp_headline}

We evaluate MSP on three multimodal benchmark datasets and compare it against state-of-the-art multimodal models. We report results for models trained with LoRA ranks $\{128, 256, 512\}$. Despite never observing real paired non-text data during training, MSP achieves competitive performance and, in several cases, outperforms baseline MLLMs on multimodal question answering tasks that require joint reasoning over paired modalities. We report audio-vision performance for all the datasets.

\begin{table}[t]
\small
\centering
\caption{Performance comparison across different multimodal models on the MUSIC-AVQA test split. Primarily, we wanted to compare against models that were allowed to use paired data but did not specifically finetune for performance on the listed benchmarks.}
\label{tab:mavqa-performance}
\begin{tabular}{lcc}
\toprule
\textbf{Method} & \textbf{Parameters} & \textbf{Performance} \\
\midrule
\midrule
ImageBind-LLM \citep{han2023imagebindllm} & 7B & 39.72 \\
X-InstructBLIP \citep{panagopoulou2023x-instructblip} & 7B & 28.10 \\
X-InstructBLIP \citep{panagopoulou2023x-instructblip} & 13B & 44.50 \\
ChatBridge \citep{zhao2023chatbridge} & 13B & 43.00 \\
OneLLM \citep{han2024onellm} & 7B & 47.60 \\
Qwen2.5-VL \citep{Qwen2.5-VL} & 7B & 47.75 \\
GPT-4o \citep{hurst2024gpt40} & - & 50.08 \\
\midrule
MSP (LoRA r = 128) & 8B & \textbf{60.24} \\
MSP (LoRA r = 256) & 8B & \textbf{64.17} \\
MSP (LoRA r = 512) & 8B & \textbf{65.65} \\
\bottomrule
\end{tabular}
\end{table}
\begin{table}[t]
\small
\centering
\caption{Performance comparison across different multimodal models on the AVQA test split. Primarily, we wanted to compare against models that were allowed to use paired data but did not specifically finetune for performance on the listed benchmarks.}
\label{tab:avqa_performance}
\begin{tabular}{lcc}
\toprule
\textbf{Method} & \textbf{Parameters} & \textbf{Performance} \\
\midrule
\midrule
PSAC \citep{li2019psac} & - & 78.60 \\
ACRTransformer \citep{zhang2020actransformer} & - & 81.70 \\
HME \citep{fan2019hme} & - & 81.80 \\
LADNet \citep{li2019ladnet} & - & 81.90 \\
SSAM \citep{reza2026ssam} & 7B & 82.12 \\
HGA \citep{jiang2020hga} & - & 82.20 \\
HCRN \citep{le2020hcrn} & - & 82.50 \\
\midrule
ImageBind-LLM \citep{han2023imagebindllm} & 7B & 55.00 \\
X-InstructBLIP \citep{panagopoulou2023x-instructblip} & 7B & 44.29 \\
\midrule
MSP (LoRA r = 128) & 8B & \textbf{85.70} \\
MSP (LoRA r = 256) & 8B & \textbf{85.73} \\
MSP (LoRA r = 512) & 8B & \textbf{85.58} \\
\bottomrule
\end{tabular}
\end{table}
\begin{table}[t]
\small
\centering
\caption{Performance comparison across different multimodal models on OmniBench test split. Primarily, we wanted to compare against models that were allowed to use paired data but did not specifically finetune for performance on the listed benchmarks.}
\label{tab:omnibench-performance}
\begin{tabular}{lcc}
\toprule
\textbf{Method} & \textbf{Parameters} & \textbf{Performance} \\
\midrule
\midrule
AnyGPT \citep{zhan2024anygpt} & 7B & 18.04 \\
UnifiedIO2-large \citep{lu2024unified}  & 1.1B & 27.06 \\
Baichuan-Omni \citep{li2024baichuanomni} & 7B & 32.20 \\
VITA \citep{fu2024vita} & 8$\times$7B & 33.10 \\
VITA-1.5 \citep{fu2025vita15} & 7B & 33.40 \\
UnifiedIO2-xxlarge \citep{lu2024unified}  & 6.8B & 33.98 \\
video-SALMONN \citep{sun2024videosalmonn} & 13B & 35.64 \\
OpenOmni \citep{luo2025openomni} & 7B & 37.40 \\
UnifiedIO2-xlarge \citep{lu2024unified} & 3.2B & 38.00 \\
MiniCPM-o 2.6 \citep{yao2024minicpm} & 7B & 40.50 \\
Baichuan-Omni-1.5 \citep{baichuan2025omni15} & 7B & 42.90 \\
Gemini-1.5-Pro \citep{team2024gemini1.5} & - & 42.91 \\
Qwen2.5-Omni \citep{qwen2025omni} & 7B & \textbf{56.13} \\
\midrule
MSP (LoRA r = 128) & 8B &  42.12 \\
MSP (LoRA r = 256) & 8B &  \textbf{43.26} \\
MSP (LoRA r = 512) & 8B &  41.07 \\
\bottomrule
\end{tabular}
\end{table}

\begin{table}[t]
\small
\centering
\caption{Proxy and bridge help improve model performance. Every cell is at $r{=}512$ on the test splits.}
\label{tab:training_methods_comparison}
\small
\setlength{\tabcolsep}{8pt}
\begin{tabular}{@{}l ccc ccc@{}}
\toprule
\multirow{2}{*}{\textbf{Method}} & \multicolumn{3}{c}{\textbf{AVQA}} & \multicolumn{3}{c}{\textbf{MUSIC-AVQA}} \\
\cmidrule(lr){2-4}\cmidrule(lr){5-7}
& \textbf{Audio} & \textbf{Video} & \textbf{Audio-Video} & \textbf{Audio} & \textbf{Video} & \textbf{Audio-Video} \\
\midrule
\textcolor{red}{Without proxy and bridge} & 59.55 & 63.91 & 70.21 & 44.03 & 44.64 & 45.75 \\
\textcolor{ForestGreen}{With proxy} \textcolor{red}{without bridge} & 78.53 & 79.50 & 82.91 & 45.85 & 54.49 & 58.60\\
\textcolor{ForestGreen}{With proxy and bridge}    & \textbf{80.24} & \textbf{81.33} & \textbf{85.58} & \textbf{57.26} & \textbf{57.11} & \textbf{65.65} \\
\bottomrule
\end{tabular}
\end{table}

\textbf{Performance on MUSIC-AVQA benchmark.}
Table~\ref{tab:mavqa-performance} reports zero-shot accuracy on the MUSIC-AVQA benchmark. MSP with LoRA rank $r{=}512$ achieves $65.65$, outperforming OneLLM by $+18.05$\,pp and GPT-4o by $+15.57$\,pp, despite GPT-4o having a substantially larger parameter count. The $r{=}256$ and $r{=}128$ configurations achieve $64.17$ and $60.24$, respectively, both surpassing the baseline methods. Unlike OneLLM, which is trained only on modality-text pairs, MSP additionally uses proxy textual descriptions of unseen modalities during training. This enables effective reasoning over paired audio-visual inputs at inference time and demonstrates the effectiveness of proxy supervision for learning cross-modal composition.

\textbf{Performance on AVQA.}
Table~\ref{tab:avqa_performance} reports performance on the AVQA  benchmark, where MSP achieves $85.73$ in zero-shot setting with LoRA rank $r{=}256$, representing the highest reported zero-shot AVQA performance among methods not finetuned on the AVQA dataset. Performance remains consistent across ranks $r{\in}\{128,256,512\}$, suggesting performance saturation rather than insufficient model capacity. MSP outperforms zero-shot baselines such as ImageBind-LLM and X-InstructBLIP by large margins ($30.73$--$41.44$\,pp). Notably, MSP also surpasses methods including PSAC, HGA, and HCRN, despite those models being trained on the paired AVQA train set. These results highlight the effectiveness of MSP in learning cross-modal composition without real paired non-text training data.

\textbf{Performance on OmniBench.}
Table~\ref{tab:omnibench-performance} reports performance on the OmniBench benchmark, which requires joint reasoning over vision, audio, and text modalities. MSP with LoRA rank $r{=}256$ achieves $43.26$, outperforming most baseline methods and achieving performance competitive with Baichuan-Omni-1.5 and Gemini-1.5-Pro despite never being trained on real paired non-text data. Notably, the closed-source Gemini-1.5-Pro achieves $42.91$, slightly below MSP on this benchmark. MSP is the only method in this comparison trained without real paired cross-modal data, and it remains competitive with models that use it, further demonstrating the effectiveness of proxy-based cross-modal composition.

\subsection{Learning Modality Composition via Proxy}
\label{sec:proxy_contribution}
We analyze the model to understand how proxy supervision and the text-to-modality bridge contribute to learning cross-modal composition. Table~\ref{tab:training_methods_comparison} compares no proxies and no bridge, proxies without a bridge, and full MSP with both proxies and bridge. At the audio-video setting, proxies are worth $+12.70$ on AVQA and $+12.85$ on MUSIC-AVQA, and the bridge a further $+2.67$ and $+7.05$. This shows that proxies provide strong learning signals for modality composition, while the bridge further improves alignment.

We also examine representation behavior across LLM layers in Table~\ref{tab:logit_lens} using both residual-stream similarity and the \emph{logit lens}. While residual similarity decreases across layers, vocabulary-space similarity recovers at the final layer and ends well above it ($\Delta=+0.166$ audio, $+0.124$ vision). This indicates that proxies do not need to match real modality features in hidden space; instead, the model learns to map them to similar output predictions, enabling effective cross-modal composition.
\begin{table}[htbp]
\scriptsize
\centering
\caption{{Paired residual and output-layer cosine similarity.}
Each proxy is matched to its nearest real counterpart by last-layer residual
cosine, and similarities are computed over those pairs. Here,
$\Delta=\text{vocab cos}_{L_{\text{last}}}-\text{residual cos}_{L_{\text{last}}}$.
These are not comparable to the trajectory cosines in
Table~\ref{tab:slot_ablation}, which measure a different quantity.}
\label{tab:logit_lens}
\small
\setlength{\tabcolsep}{6pt}
\renewcommand{\arraystretch}{1.10}
\begin{tabular}{@{}l c c c@{}}
\toprule
Modality & residual cos ($L_0/L_{\text{mid}}/L_{\text{last}}$)
         & vocab cos ($L_0/L_{\text{mid}}/L_{\text{last}}$)
         & $\Delta$ at last layer \\
\midrule
Audio    & $.956\,/\,.889\,/\,.778$ & $.953\,/\,.913\,/\,.944$ & $\mathbf{+0.166}$ \\
Vision   & $.821\,/\,.883\,/\,.805$ & $.828\,/\,.799\,/\,.929$ & $\mathbf{+0.124}$ \\
\bottomrule
\end{tabular}
\end{table}

We fit a logistic-regression probe at the input, middle, and final layers of the LLM backbone to predict real vs.\ bridged-proxy from the hidden state at that layer ($n{=}400$ per cell). Real and bridged-proxy features remain linearly separable at every layer for both audio and vision (Table~\ref{tab:linear_probe}); MSP does not erase the distinction in the residual stream. This supports the interpretation that the LM-head readout, rather than residual indistinguishability, explains the proxy--real alignment.
\begin{table}[htbp]
\centering
\caption{{Real and proxy hidden states remain distinguishable.}
A logistic-regression probe predicts whether a hidden state came from
a real feature or a bridged-proxy feature at three layers; $n{=}400$
per cell. Values are probe accuracies, so higher values mean real and
proxy states are easier to separate. The high accuracies show that MSP
does not make the hidden states identical; Table~\ref{tab:logit_lens}
tests whether the LM-head readout is nevertheless aligned.}
\label{tab:linear_probe}
\small
\setlength{\tabcolsep}{12pt}
\renewcommand{\arraystretch}{1.10}
\begin{tabular}{@{}l c c c@{}}
\toprule
Modality      & $L_0$  & $L_{\text{mid}}$ & $L_{\text{last}}$ \\
\midrule
Audio         & 0.953  & 0.970            & 0.963 \\
Vision        & 1.000  & 1.000            & 0.998 \\
\bottomrule
\end{tabular}\\[2pt]
\end{table}

Additional alignment analyses are provided in Appendix~\ref{app:mechanism_probes}.

Phase~2 trains on QA-formatted data, so part of the gain over Phase~1 comes from that supervision alone rather than from composition. Separating the two (Table~\ref{tab:training_stages}), we still find that MSP composition training with proxies enables significantly better performance than the QA supervision by itself.

\begin{table}[t]
\centering
\small
\caption{Real modalities vs. proxy modalities at inference and token-matched controls. Qwen3-8B LoRA $r{=}512$, test splits: AVQA ($n{=}16{,}910$), MUSIC-AVQA ($n{=}9{,}185$).}
\label{tab:rebuttal_controls}
\setlength{\tabcolsep}{5pt}
\begin{tabular*}{\linewidth}{@{\extracolsep{\fill}}llrr@{}}
\toprule
\multicolumn{4}{@{}l}{\textbf{(a) Real modalities vs. proxy modalities at inference}} \\
Audio & Vision & AVQA & MUSIC-AVQA \\
\midrule
real & real & 85.58 & 65.65 \\
imagined & real & 81.19 & 57.34 \\
real & imagined & 79.28 & 62.87 \\
imagined & imagined & 75.75 & 41.33 \\
\bottomrule
\end{tabular*}
\par\smallskip
\begin{tabular*}{\linewidth}{@{\extracolsep{\fill}}lrrrr@{}}
\toprule
\multicolumn{5}{@{}l}{\textbf{(b) Token-matched controls}} \\
& \multicolumn{2}{c}{Audio slot} & \multicolumn{2}{c}{Vision slot} \\
\cmidrule(lr){2-3}\cmidrule(l){4-5}
& AVQA & MUSIC-AVQA & AVQA & MUSIC-AVQA \\
\midrule
Real & 85.58 & 65.65 & 85.58 & 65.65 \\
Learned null & 81.50 & 60.48 & 80.24 & 61.35 \\
Zero & 80.28 & 56.76 & 78.35 & 59.01 \\
Random (moment-matched Gaussian) & 79.66 & 56.08 & 74.97 & 58.08 \\
Shuffled (real feature, wrong example) & 76.81 & 52.95 & 72.89 & 54.48 \\
\bottomrule
\end{tabular*}
\end{table}

\textbf{Real modalities vs. proxy modalities at inference.}
We found across the board that real modalities outperform their proxy counterparts (Table~\ref{tab:rebuttal_controls}a). Token count, position, attention mask, and decoding are unchanged. Proxies for this ablation are generated via the same mechanism used during the training of the model.

\textbf{Token-matched controls.}
Table~\ref{tab:rebuttal_controls}b varies the audio slot with real vision and the vision slot with real audio.
The key takeaway is that the model utilizes real modality content through these slots, and that only correct content recovers full performance. Additional details are provided in Appendix~\ref{app:rebuttal_controls}.
Question-type breakdowns at $r{=}512$ are provided in Appendix~\ref{app:qtype_breakdown}.

\textbf{Sensitivity to proxy quality.}
As might be expected based on the token-matched controls, the quality and content of the modality given to the model has a large impact. While we do not have results for a model trained on degenerate proxies, the token-matched controls indicate that the model is reading real modality content from the token slot, thus generating realistic proxies is also important for the downstream model to learn the desired cross-modal relationships (Appendix~\ref{app:proxy_quality}).

MUSIC-AVQA-R results are provided in Appendix~\ref{app:musicavqar}.

\section{Limitations and Future Directions}
\label{sec:limitations}

While MSP demonstrates that proxy-mediated alignment can enable multimodal composition without real paired non-text supervision, it relies on the quality and semantic consistency of proxy textual descriptions. Noisy or weakly aligned proxies can degrade model performance (Appendix~\ref{app:proxy_quality}).
Additionally, MSP builds on contrastively pretrained encoders (e.g., CLIP/CLAP-style models) to provide cross-modal structure, and therefore inherits their coverage limitations and potential biases.
Future work can address these limitations along several directions. Improving proxy generation—through stronger generative models or retrieval-based methods—could yield more accurate and semantically grounded descriptions. Reducing the training–inference gap, either by better integrating proxy information at inference or by learning proxy-free alignment mechanisms, is another promising avenue. Extending MSP to a wider range of modalities and more complex reasoning tasks, as well as studying its behavior at larger scales, would further evaluate its generality. Finally, developing a deeper theoretical understanding 
may help clarify when and why multimodal composition can emerge without real paired non-text supervision. Future work should focus on using MSP in paradigms with significantly more tokens and also determine how closely MSP compares with paired data ceilings (or if they can be exceeded by using additional text proxies; Appendix~\ref{app:paired_pilot}).

\textbf{Broader impacts.}
This work improves multimodal reasoning without requiring real paired non-text data, which can benefit data-limited or privacy-sensitive applications. However, improved cross-modal synthesis may also facilitate misuse such as multimodal deepfakes or fabricated content. To mitigate these dual-use risks, we advocate for the use of ethically sourced training data, model access control, and the adoption of provenance-tracking techniques for any synthesized outputs.

\section{Conclusion}
\label{sec:conclusion}
In this work, we revisited a central assumption in multimodal learning that large-scale paired multimodal data is necessary for learning cross-modal reasoning. We introduced \textbf{Modality Self-Play (MSP)}, a training paradigm in which proxy textual descriptions serve as semantic bridges between observed and unseen modalities. MSP removes the need for real paired non-text (e.g., audio-vision) supervision, while continuing to rely on modality-text pairs, pretrained contrastive encoders, and modality-distribution bridge calibration.
Our results show that proxy-based supervision is both viable and effective. Models trained with MSP demonstrate strong performance on standard multimodal benchmarks, while reducing reliance on costly data curation pipelines. 
More broadly, our findings suggest that multimodal composition can emerge from structured interactions over independently learned representations mediated through proxy text, rather than from exhaustive aligned multimodal data. We hope this work encourages the development of more scalable and data-efficient approaches to multimodal learning based on principled alignment mechanisms.

\section*{Acknowledgments}
This work is partially supported by a gift award from Amazon and an NSF award 2406199.

\bibliographystyle{plain}
\bibliography{main}

@inproceedings{wu2023multimodalSurvey,
  title={Multimodal large language models: A survey},
  author={Wu, Jiayang and Gan, Wensheng and Chen, Zefeng and Wan, Shicheng and Yu, Philip S},
  booktitle={2023 IEEE International Conference on Big Data (BigData)},
  pages={2247--2256},
  year={2023},
  organization={IEEE}
}

@article{huang2024multimodalSurvey,
  title={From large language models to large multimodal models: A literature review},
  author={Huang, Dawei and Yan, Chuan and Li, Qing and Peng, Xiaojiang},
  journal={Applied Sciences},
  volume={14},
  number={12},
  pages={5068},
  year={2024},
  publisher={MDPI}
}

@article{han2023imagebindllm,
  title={Imagebind-llm: Multi-modality instruction tuning},
  author={Han, Jiaming and Zhang, Renrui and Shao, Wenqi and Gao, Peng and Xu, Peng and Xiao, Han and Zhang, Kaipeng and Liu, Chris and Wen, Song and Guo, Ziyu and others},
  journal={arXiv preprint arXiv:2309.03905},
  year={2023}
}

@inproceedings{han2024onellm,
  title={Onellm: One framework to align all modalities with language},
  author={Han, Jiaming and Gong, Kaixiong and Zhang, Yiyuan and Wang, Jiaqi and Zhang, Kaipeng and Lin, Dahua and Qiao, Yu and Gao, Peng and Yue, Xiangyu},
  booktitle={Proceedings of the IEEE/CVF Conference on Computer Vision and Pattern Recognition},
  pages={26584--26595},
  year={2024}
}

@article{panagopoulou2023x-instructblip,
  title={X-instructblip: A framework for aligning x-modal instruction-aware representations to llms and emergent cross-modal reasoning},
  author={Panagopoulou, Artemis and Xue, Le and Yu, Ning and Li, Junnan and Li, Dongxu and Joty, Shafiq and Xu, Ran and Savarese, Silvio and Xiong, Caiming and Niebles, Juan Carlos},
  journal={arXiv preprint arXiv:2311.18799},
  year={2023}
}

@inproceedings{wu2024nextgpt,
  title={Next-gpt: Any-to-any multimodal llm},
  author={Wu, Shengqiong and Fei, Hao and Qu, Leigang and Ji, Wei and Chua, Tat-Seng},
  booktitle={Forty-first International Conference on Machine Learning},
  year={2024}
}

@inproceedings{zhan2024anygpt,
  title={Anygpt: Unified multimodal llm with discrete sequence modeling},
  author={Zhan, Jun and Dai, Junqi and Ye, Jiasheng and Zhou, Yunhua and Zhang, Dong and Liu, Zhigeng and Zhang, Xin and Yuan, Ruibin and Zhang, Ge and Li, Linyang and others},
  booktitle={Proceedings of the 62nd Annual Meeting of the Association for Computational Linguistics (Volume 1: Long Papers)},
  pages={9637--9662},
  year={2024}
}

@inproceedings{su2023pandagpt,
  title={Pandagpt: One model to instruction-follow them all},
  author={Su, Yixuan and Lan, Tian and Li, Huayang and Xu, Jialu and Wang, Yan and Cai, Deng},
  booktitle={Proceedings of the 1st Workshop on Taming Large Language Models: Controllability in the era of Interactive Assistants!},
  pages={11--23},
  year={2023}
}

@inproceedings{radford2021clip,
  title={Learning transferable visual models from natural language supervision},
  author={Radford, Alec and Kim, Jong Wook and Hallacy, Chris and Ramesh, Aditya and Goh, Gabriel and Agarwal, Sandhini and Sastry, Girish and Askell, Amanda and Mishkin, Pamela and Clark, Jack and others},
  booktitle={International conference on machine learning},
  pages={8748--8763},
  year={2021},
  organization={PmLR}
}

@inproceedings{wu2023clap,
  title={Large-scale contrastive language-audio pretraining with feature fusion and keyword-to-caption augmentation},
  author={Wu, Yusong and Chen, Ke and Zhang, Tianyu and Hui, Yuchen and Berg-Kirkpatrick, Taylor and Dubnov, Shlomo},
  booktitle={ICASSP 2023-2023 IEEE International Conference on Acoustics, Speech and Signal Processing (ICASSP)},
  pages={1--5},
  year={2023},
  organization={IEEE}
}

@inproceedings{li2023blip2,
  title={{BLIP-2}: Bootstrapping language-image pre-training with frozen image encoders and large language models},
  author={Li, Junnan and Li, Dongxu and Savarese, Silvio and Hoi, Steven},
  booktitle={Proceedings of the 40th International Conference on Machine Learning (ICML)},
  year={2023}
}

@inproceedings{liu2024llava15,
  title={Improved baselines with visual instruction tuning},
  author={Liu, Haotian and Li, Chunyuan and Li, Yuheng and Lee, Yong Jae},
  booktitle={Proceedings of the IEEE/CVF Conference on Computer Vision and Pattern Recognition (CVPR)},
  pages={26296--26306},
  year={2024}
}

@inproceedings{liu2023llava,
  title={Visual instruction tuning},
  author={Liu, Haotian and Li, Chunyuan and Wu, Qingyang and Lee, Yong Jae},
  booktitle={Advances in Neural Information Processing Systems (NeurIPS)},
  year={2023}
}

@article{wang2024mio,
  title={{MIO}: A foundation model on multimodal tokens},
  author={Wang, Zekun and Zhu, King and Xu, Chunpu and Zhou, Wangchunshu and Liu, Jiaheng and Zhang, Yibo and Wang, Jiashuo and Shi, Ning and Li, Siyu and Li, Yizhi and others},
  journal={arXiv preprint arXiv:2409.17692},
  year={2024}
}

@article{qwen2025omni,
  title={{Qwen2.5-Omni} technical report},
  author={{Qwen Team}},
  journal={arXiv preprint arXiv:2503.20215},
  year={2025}
}

@article{baichuan2025omni15,
  title={{Baichuan-Omni-1.5} technical report},
  author={Li, Yadong and Liu, Jun and Zhang, Tao and Zhang, Tao and Chen, Song and Li, Tianpeng and Li, Zehuan and Liu, Lijun and Ming, Lingfeng and Dong, Guosheng and others},
  journal={arXiv preprint arXiv:2501.15368},
  year={2025}
}

@article{fu2025vita15,
  title={{VITA-1.5}: Towards {GPT-4o} level real-time vision and speech interaction},
  author={Fu, Chaoyou and Lin, Haojia and Wang, Xiong and Zhang, Yi-Fan and Shen, Yunhang and Liu, Xiaoyu and others},
  journal={arXiv preprint arXiv:2501.01957},
  year={2025}
}

@inproceedings{sun2024videosalmonn,
  title={video-{SALMONN}: Speech-enhanced audio-visual large language models},
  author={Sun, Guangzhi and Yu, Wenyi and Tang, Changli and Chen, Xianzhao and Tan, Tian and Li, Wei and Lu, Lu and Ma, Zejun and Wang, Yuxuan and Zhang, Chao},
  booktitle={Proceedings of the 41st International Conference on Machine Learning (ICML)},
  year={2024}
}

@inproceedings{luo2025openomni,
  title={{OpenOmni}: Advancing open-source omnimodal large language models with progressive multimodal alignment and real-time self-aware emotional speech synthesis},
  author={Luo, Run and Lin, Ting-En and Zhang, Haonan and Wu, Yuchuan and Liu, Xiong and Yang, Min and Li, Yongbin},
  booktitle={Advances in Neural Information Processing Systems (NeurIPS)},
  year={2025}
}

@article{zhao2023chatbridge,
  title={{ChatBridge}: Bridging modalities with large language model as a language catalyst},
  author={Zhao, Zijia and Guo, Longteng and Yue, Tongtian and Chen, Sihan and Shao, Shuai and Zhu, Xinxin and Yuan, Zehuan and Liu, Jing},
  journal={arXiv preprint arXiv:2305.16103},
  year={2023}
}

@inproceedings{li2023decap,
  title={{DeCap}: Decoding {CLIP} latents for zero-shot captioning via text-only training},
  author={Li, Wei and Zhu, Linchao and Wen, Longyin and Yang, Yi},
  booktitle={International Conference on Learning Representations (ICLR)},
  year={2023}
}

@inproceedings{nukrai2022capdec,
  title={Text-only training for image captioning using noise-injected {CLIP}},
  author={Nukrai, David and Mokady, Ron and Globerson, Amir},
  booktitle={Findings of the Association for Computational Linguistics: EMNLP 2022},
  year={2022}
}

@inproceedings{wang2023knight,
  title={From association to generation: Text-only captioning by unsupervised cross-modal mapping},
  author={Wang, Junyang and Yan, Ming and Zhang, Yi and Sang, Jitao},
  booktitle={Proceedings of the Thirty-Second International Joint Conference on Artificial Intelligence (IJCAI)},
  pages={4326--4334},
  year={2023}
}

@inproceedings{deshmukh2024noaudio,
  title={Training audio captioning models without audio},
  author={Deshmukh, Soham and Elizalde, Benjamin and Emmanouilidou, Dimitra and Raj, Bhiksha and Singh, Rita and Wang, Huaming},
  booktitle={IEEE International Conference on Acoustics, Speech and Signal Processing (ICASSP)},
  year={2024}
}

@inproceedings{li2025drcap,
  title={{DRCap}: Decoding {CLAP} latents with retrieval-augmented generation for zero-shot audio captioning},
  author={Li, Xiquan and Chen, Wenxi and Ma, Ziyang and Xu, Xuenan and Liang, Yuzhe and Zheng, Zhisheng and Kong, Qiuqiang and Chen, Xie},
  booktitle={IEEE International Conference on Acoustics, Speech and Signal Processing (ICASSP)},
  year={2025}
}

@inproceedings{xu2024metaclip,
  title={Demystifying {CLIP} data},
  author={Xu, Hu and Xie, Saining and Tan, Xiaoqing Ellen and Huang, Po-Yao and Howes, Russell and Sharma, Vasu and Li, Shang-Wen and Ghosh, Gargi and Zettlemoyer, Luke and Feichtenhofer, Christoph},
  booktitle={International Conference on Learning Representations (ICLR)},
  year={2024}
}

@inproceedings{girdhar2023imagebind,
  title={{ImageBind}: One embedding space to bind them all},
  author={Girdhar, Rohit and El-Nouby, Alaaeldin and Liu, Zhuang and Singh, Mannat and Alwala, Kalyan Vasudev and Joulin, Armand and Misra, Ishan},
  booktitle={Proceedings of the IEEE/CVF Conference on Computer Vision and Pattern Recognition (CVPR)},
  year={2023}
}

@inproceedings{liang2022mindthegap,
  title={Mind the gap: Understanding the modality gap in multi-modal contrastive representation learning},
  author={Liang, Victor Weixin and Zhang, Yuhui and Kwon, Yongchan and Yeung, Serena and Zou, James Y},
  booktitle={Advances in Neural Information Processing Systems (NeurIPS)},
  year={2022}
}

@article{qwen2025qwen3,
  title={{Qwen3} technical report},
  author={{Qwen Team}},
  journal={arXiv preprint arXiv:2505.09388},
  year={2025}
}

@inproceedings{li2022musicavqa,
  title={Learning to answer questions in dynamic audio-visual scenarios},
  author={Li, Guangyao and Wei, Yake and Tian, Yapeng and Xu, Chenliang and Wen, Ji-Rong and Hu, Di},
  booktitle={Proceedings of the IEEE/CVF Conference on Computer Vision and Pattern Recognition (CVPR)},
  year={2022}
}

@article{team2024gemini1.5,
  title={Gemini 1.5: Unlocking multimodal understanding across millions of tokens of context},
  author={Team, Gemini and Georgiev, Petko and Lei, Ving Ian and Burnell, Ryan and Bai, Libin and Gulati, Anmol and Tanzer, Garrett and Vincent, Damien and Pan, Zhufeng and Wang, Shibo and others},
  journal={arXiv preprint arXiv:2403.05530},
  year={2024}
}

@article{yao2024minicpm,
  title={MiniCPM-V: A GPT-4V Level MLLM on Your Phone},
  author={Yao, Yuan and Yu, Tianyu and Zhang, Ao and Wang, Chongyi and Cui, Junbo and Zhu, Hongji and Cai, Tianchi and Li, Haoyu and Zhao, Weilin and He, Zhihui and others},
  journal={arXiv preprint arXiv:2408.01800},
  year={2024}
}

@article{fu2024vita,
  title={Vita: Towards open-source interactive omni multimodal llm},
  author={Fu, Chaoyou and Lin, Haojia and Long, Zuwei and Shen, Yunhang and Dai, Yuhang and Zhao, Meng and Zhang, Yi-Fan and Dong, Shaoqi and Li, Yangze and Wang, Xiong and others},
  journal={arXiv preprint arXiv:2408.05211},
  year={2024}
}

@inproceedings{lu2024unified,
  title={Unified-io 2: Scaling autoregressive multimodal models with vision language audio and action},
  author={Lu, Jiasen and Clark, Christopher and Lee, Sangho and Zhang, Zichen and Khosla, Savya and Marten, Ryan and Hoiem, Derek and Kembhavi, Aniruddha},
  booktitle={Proceedings of the IEEE/CVF Conference on Computer Vision and Pattern Recognition},
  pages={26439--26455},
  year={2024}
}

@article{hurst2024gpt40,
  title={Gpt-4o system card},
  author={Hurst, Aaron and Lerer, Adam and Goucher, Adam P and Perelman, Adam and Ramesh, Aditya and Clark, Aidan and Ostrow, AJ and Welihinda, Akila and Hayes, Alan and Radford, Alec and others},
  journal={arXiv preprint arXiv:2410.21276},
  year={2024}
}

@article{Qwen2.5-VL,
  title={Qwen2.5-VL Technical Report},
  author={Bai, Shuai and Chen, Keqin and Liu, Xuejing and Wang, Jialin and Ge, Wenbin and Song, Sibo and Dang, Kai and Wang, Peng and Wang, Shijie and Tang, Jun and Zhong, Humen and Zhu, Yuanzhi and Yang, Mingkun and Li, Zhaohai and Wan, Jianqiang and Wang, Pengfei and Ding, Wei and Fu, Zheren and Xu, Yiheng and Ye, Jiabo and Zhang, Xi and Xie, Tianbao and Cheng, Zesen and Zhang, Hang and Yang, Zhibo and Xu, Haiyang and Lin, Junyang},
  journal={arXiv preprint arXiv:2502.13923},
  year={2025}
}

@inproceedings{fan2019hme,
  title={Heterogeneous memory enhanced multimodal attention model for video question answering},
  author={Fan, Chenyou and Zhang, Xiaofan and Zhang, Shu and Wang, Wensheng and Zhang, Chi and Huang, Heng},
  booktitle={Proceedings of the IEEE/CVF conference on computer vision and pattern recognition},
  pages={1999--2007},
  year={2019}
}

@inproceedings{jiang2020hga,
  title={Reasoning with heterogeneous graph alignment for video question answering},
  author={Jiang, Pin and Han, Yahong},
  booktitle={Proceedings of the AAAI conference on artificial intelligence},
  volume={34},
  pages={11109--11116},
  year={2020}
}

@inproceedings{le2020hcrn,
  title={Hierarchical conditional relation networks for video question answering},
  author={Le, Thao Minh and Le, Vuong and Venkatesh, Svetha and Tran, Truyen},
  booktitle={Proceedings of the IEEE/CVF conference on computer vision and pattern recognition},
  pages={9972--9981},
  year={2020}
}

@inproceedings{li2019ladnet,
  title={Learnable aggregating net with diversity learning for video question answering},
  author={Li, Xiangpeng and Gao, Lianli and Wang, Xuanhan and Liu, Wu and Xu, Xing and Shen, Heng Tao and Song, Jingkuan},
  booktitle={Proceedings of the 27th ACM international conference on multimedia},
  pages={1166--1174},
  year={2019}
}

@inproceedings{li2019psac,
  title={Beyond rnns: Positional self-attention with co-attention for video question answering},
  author={Li, Xiangpeng and Song, Jingkuan and Gao, Lianli and Liu, Xianglong and Huang, Wenbing and He, Xiangnan and Gan, Chuang},
  booktitle={Proceedings of the AAAI conference on artificial intelligence},
  volume={33},
  pages={8658--8665},
  year={2019}
}

@article{zhang2020actransformer,
  title={Action-centric relation transformer network for video question answering},
  author={Zhang, Jipeng and Shao, Jie and Cao, Rui and Gao, Lianli and Xu, Xing and Shen, Heng Tao},
  journal={IEEE Transactions on Circuits and Systems for Video Technology},
  volume={32},
  number={1},
  pages={63--74},
  year={2020},
  publisher={IEEE}
}

@article{reza2026ssam,
  title={SSAM: Singular Subspace Alignment for Merging Multimodal Large Language Models},
  author={Reza, Md Kaykobad and Patil, Ameya and Ayrapetian, Edward and Asif, M Salman},
  journal={arXiv preprint arXiv:2603.21584},
  year={2026}
}

@inproceedings{yang2022avqa,
  title={{AVQA}: A Dataset for Audio-Visual Question Answering on Videos},
  author={Yang, Pinci and Wang, Xin and Duan, Xuguang and Chen, Hong and Hou, Runze and Jin, Cong and Zhu, Wenwu},
  booktitle={Proceedings of the 30th ACM International Conference on Multimedia (MM)},
  year={2022}
}

@inproceedings{audiocaps,
  title={AudioCaps: Generating Captions for Audios in The Wild},
  author={Kim, Chris Dongjoo and Kim, Byeongchang and Lee, Hyunmin and Kim, Gunhee},
  booktitle={NAACL-HLT},
  year={2019}
}

@article{mei2024wavcaps,
  title={Wavcaps: A chatgpt-assisted weakly-labelled audio captioning dataset for audio-language multimodal research},
  author={Mei, Xinhao and Meng, Chutong and Liu, Haohe and Kong, Qiuqiang and Ko, Tom and Zhao, Chengqi and Plumbley, Mark D and Zou, Yuexian and Wang, Wenwu},
  journal={IEEE/ACM Transactions on Audio, Speech, and Language Processing},
  volume={32},
  pages={3339--3354},
  year={2024},
  publisher={IEEE}
}

@inproceedings{lipping2022clotho,
  title={{Clotho-AQA}: A crowdsourced dataset for audio question answering},
  author={Lipping, Samuel and Sudarsanam, Parthasaarathy and Drossos, Konstantinos and Virtanen, Tuomas},
  booktitle={2022 30th European Signal Processing Conference (EUSIPCO)},
  pages={1140--1144},
  year={2022},
  organization={IEEE}
}

@article{li2024omnibench,
  title={Omnibench: Towards the future of universal omni-language models},
  author={Li, Yizhi and Ma, Yinghao and Zhang, Ge and Yuan, Ruibin and Zhu, Kang and Guo, Hangyu and Liang, Yiming and Liu, Jiaheng and Wang, Zekun and Yang, Jian and others},
  journal={arXiv preprint arXiv:2409.15272},
  year={2024}
}

@inproceedings{xu2017videoqa,
  title={Video question answering via gradually refined attention over appearance and motion},
  author={Xu, Dejing and Zhao, Zhou and Xiao, Jun and Wu, Fei and Zhang, Hanwang and He, Xiangnan and Zhuang, Yueting},
  booktitle={Proceedings of the 25th ACM International Conference on Multimedia (MM)},
  pages={1645--1653},
  year={2017}
}

@inproceedings{goyal2017vqav2,
  title={Making the {V} in {VQA} matter: Elevating the role of image understanding in visual question answering},
  author={Goyal, Yash and Khot, Tejas and Summers-Stay, Douglas and Batra, Dhruv and Parikh, Devi},
  booktitle={Proceedings of the IEEE Conference on Computer Vision and Pattern Recognition (CVPR)},
  pages={6904--6913},
  year={2017}
}

@inproceedings{ma2024musicavqar,
  title={Look, Listen, and Answer: Overcoming Biases for Audio-Visual Question Answering},
  author={Ma, Jie and Hu, Min and Wang, Pinghui and Sun, Wangchun and Song, Lingyun and Pei, Hongbin and Liu, Jun and Du, Youtian},
  booktitle={Advances in Neural Information Processing Systems},
  volume={37},
  year={2024},
  url={https://proceedings.neurips.cc/paper_files/paper/2024/hash/123a18dfd821c8b440f42a00a27648d6-Abstract-Conference.html}
}

@article{li2024baichuanomni,
  title={{Baichuan-Omni} technical report},
  author={Li, Yadong and Sun, Haoze and Lin, Mingan and Li, Tianpeng and
          Dong, Guosheng and Zhang, Tao and Ding, Bowen and Song, Wei and
          Cheng, Zhenglin and Huo, Yuqi and others},
  journal={arXiv preprint arXiv:2410.08565},
  year={2024}
}

\clearpage
\appendix
\raggedbottom
\section{Detailed architecture diagram}
\label{app:architecture_detail}

\begin{figure}[htbp]
  \centering
  \includegraphics[width=\linewidth]{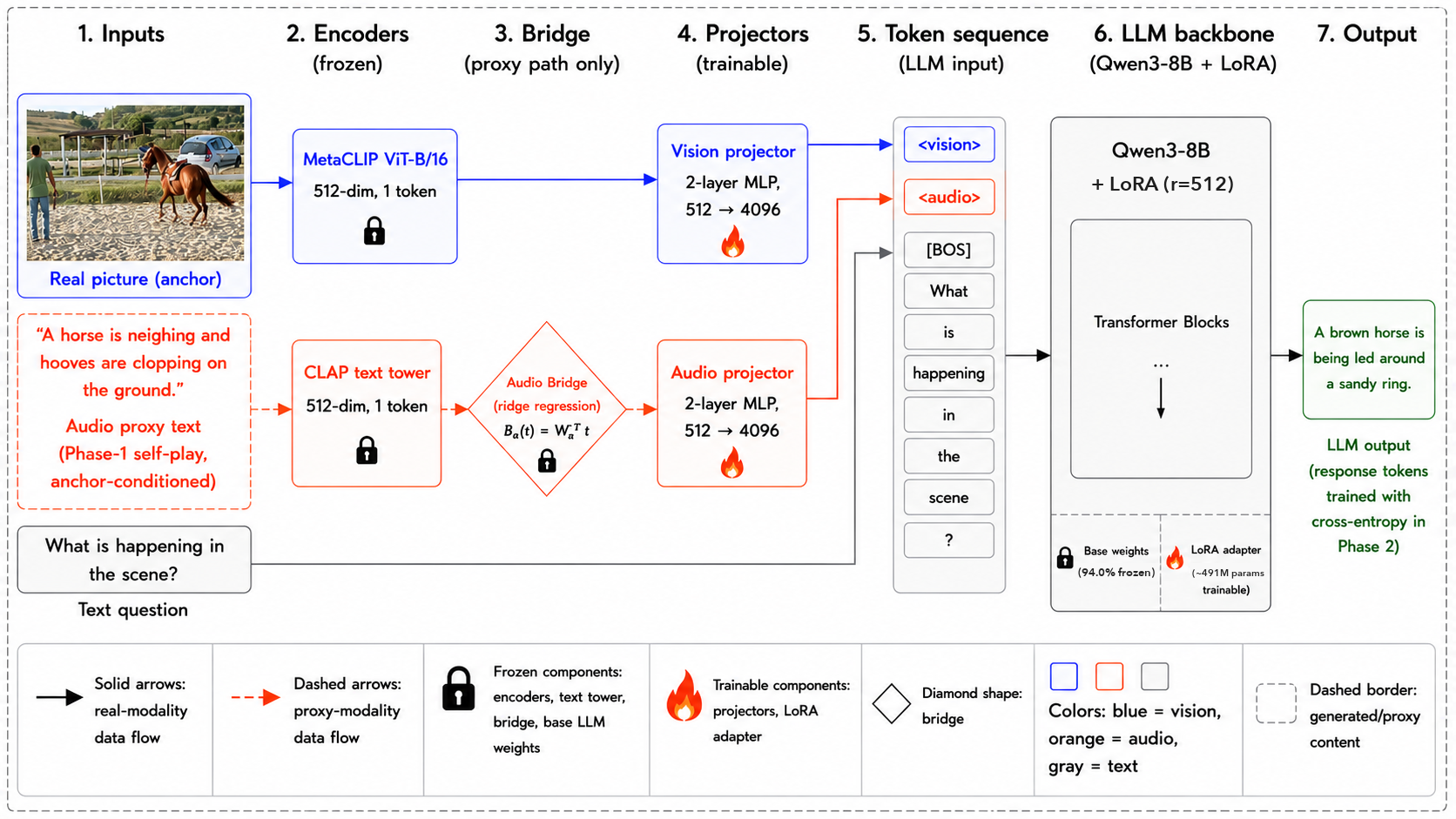}
  \caption{{MSP architecture, component-level view.} MSP uses frozen audio and
  vision encoders with their associated text encoders, trainable per-modality
  projectors, a frozen text-to-modality bridge for Phase~2, and a Qwen3-8B
  language model with LoRA adapters.}
  \label{fig:msp_architecture_detail}
\end{figure}

Figure~\ref{fig:msp_architecture_detail} gives the component-level view of
MSP referred to in Section~\ref{sec:model-architecture}. It shows the frozen
contrastive encoders and their text towers, the trainable per-modality
projectors, the frozen text-to-modality bridge used on the proxy path in
Phase~2, the reserved token slots in the prompt, and the Qwen3-8B backbone
with its LoRA adapters.

\FloatBarrier

\section{Per-question-type breakdowns}
\label{app:qtype_breakdown}

The single-token model does not collapse: AVQA \emph{Come From} (sound localization) $90.9$ and \emph{Where} (spatial) $80.3$; MUSIC-AVQA \emph{Counting} $72.6$ and \emph{Location} $57.2$ (recomputed using the $r{=}512$ checkpoint; Table~\ref{tab:qtype_r512}).

\begin{table}[htbp]
\centering
\small
\caption{Reported question-type and overall modality results at $r{=}512$. The question-type block reports the four categories recomputed for this checkpoint.}
\label{tab:qtype_r512}
\begin{tabular}{@{}llr@{}}
\toprule
Dataset & Question type & Accuracy \\
\midrule
AVQA & Come From (sound localization) & 90.9 \\
AVQA & Where (spatial) & 80.3 \\
MUSIC-AVQA & Counting & 72.6 \\
MUSIC-AVQA & Location & 57.2 \\
\bottomrule
\end{tabular}
\par\smallskip
\begin{tabular}{@{}lrrr@{}}
\toprule
Dataset & Audio & Video & Audio-Video \\
\midrule
AVQA & 80.24 & 81.33 & 85.58 \\
MUSIC-AVQA & 57.26 & 57.11 & 65.65 \\
\bottomrule
\end{tabular}
\end{table}

\FloatBarrier

\section{Self-play prompt templates}
\label{app:proxy_prompts}

Phase~2 proxies are generated by the Phase-1 checkpoint conditioned on
the real anchor modality. The prompt template controls which target
encoder distribution the generated text is meant to match; we use
encoder-native terse prompts that mirror the caption style each
contrastive encoder was trained on (Appendix \ref{app:bridge_variants}). The
verbatim templates used throughout this paper are reproduced below.

\noindent\begin{minipage}{\linewidth}
\paragraph{Audio proxy from a vision anchor (CLAP-targeting).}
\begin{quote}
\texttt{<vision>}\\[2pt]
Describe the sounds in this scene as a single AudioCaps-style
caption: ONE short sentence, 8--15 words, present tense, naming
1--3 specific sound sources directly. Do not use multiple sentences
and do not describe pitch, timbre, or volume.\\
\\
Do NOT use: image, picture, photo, visible, scene, look, see, shown,
view, visually.\\
\\
Examples:\\
\hspace*{1em} ``A man is speaking while water is flowing.''\\
\hspace*{1em} ``A train engine is running and blowing its horn.''\\
\hspace*{1em} ``Music is playing as a vehicle passes by.''\\
\hspace*{1em} ``An animal is chewing loudly in the background.''\\
\hspace*{1em} ``Wind is blowing and leaves are rustling.''\\
\\
Write ONE such sentence. No other text.
\end{quote}
\end{minipage}\par

\noindent\begin{minipage}{\linewidth}
\paragraph{Vision proxy from an audio anchor (MetaCLIP-targeting).}
\begin{quote}
\texttt{<audio>}\\[2pt]
Write a single COCO-style image caption describing what the source of
the sound would look like as a photograph. Use one short sentence,
8--18 words, present tense.\\
\\
Do NOT use any of these words or concepts: sound, audio, noise, hear,
hearing, listen, volume, pitch, timbre, loud, quiet, rhythmic, echo.\\
\\
Name the main object, person, or animal and the setting. Mention
visible attributes only.\\
\\
Examples:\\
\hspace*{1em} ``A dog stands on grass with its mouth open.''\\
\hspace*{1em} ``A man sits indoors beside a running kitchen sink.''\\
\hspace*{1em} ``A train passes along tracks near a platform.''\\
\\
Write ONE such sentence. No other text.
\end{quote}
\end{minipage}\par

Both templates share a common structure: one paragraph naming the
target encoder's caption-style family, one paragraph of negative-word
constraints to prevent leakage from the source modality's vocabulary,
and a short positive instruction listing the content axes the target
caption distribution covers.

\section{Optimization}
\label{app:phase1_config}

Both Phase~1 and Phase~2 use AdamW with bfloat16 mixed precision,
peak learning rate $5{\times}10^{-5}$, cosine decay, one epoch per
phase, and seed $42$. Effective batch size is $16$ (micro-batch $2$
$\times$ gradient-accumulation $8$) for LoRA $r{=}128$ and $16$
(micro-batch $1$ $\times$ gradient-accumulation $16$) for
$r\in\{256,512\}$. Each reported LoRA run uses DeepSpeed ZeRO-2 on
one NVIDIA H100 (80\,GB).

\paragraph{Phase~1 schedule.}
For the first $2{,}000$ optimizer steps the LLM is frozen and only
the modality projectors are updated; afterward the LoRA parameters
are unfrozen. Each Phase~1 batch carries real features from exactly
one modality, and audio and vision batches alternate so the update
parameters see both pathways within each rotation. The Phase-1
checkpoint is then used both as the initialization for Phase~2 and as
the generator that produces the cached self-play proxies.

\paragraph{Phase~2 schedule.}
Phase~2 continues from the Phase-1 checkpoint with frozen bridges
and the cached self-play proxies; only the modality projectors and
the chosen LLM update parameters are trained. Dataset sources for
both phases are listed in Section~\ref{sec:exp_setup}.

One additional note is that for all models, modality dropout is applied during training so that absent modalities are not treated as malformed input.

\section{Detailed mechanism probes}
\label{app:mechanism_probes}

To analyze the effect of bridge on modality alignment, we calculate the cosine similarity between proxy and real modality features. As shown in Table~\ref{tab:cosine_alignment}, alignment improves substantially after applying the bridge, increasing similarity from $0.218$--$0.422$ in the raw
encoder space to $0.548$--$0.823$ after bridging (a $1.8\times$--$2.6\times$
increase), with additional smaller gains from the projector.

\begin{table}[htbp]
\centering
\caption{{Bridge and projector progressively align proxy with
real-modality features.}}
\label{tab:cosine_alignment}
\small
\setlength{\tabcolsep}{8pt}
\renewcommand{\arraystretch}{1.15}
\begin{tabular}{@{}l l c c c@{}}
\toprule
\multirow{2}{*}{\textbf{Dataset}}      & \multirow{2}{*}{\textbf{Modality}} & \textbf{Pre-bridge} & \textbf{Pre-projector} & \textbf{Post-projector} \\
& & (encoder only) & (encoder+bridge) & (encoder+bridge+projector) \\
\midrule
\multirow{2}{*}{AVQA}         & Vision   & 0.307      & 0.786 \textcolor{ForestGreen}{\bf \scriptsize (+0.479)}        & \textbf{0.889} \textcolor{ForestGreen}{\bf \scriptsize (+0.582)} \\
   & Audio    & 0.218      & 0.548 \textcolor{ForestGreen}{\bf \scriptsize (+0.330)}        & \textbf{0.697} \textcolor{ForestGreen}{\bf \scriptsize (+0.479)} \\
\midrule
\multirow{2}{*}{MUSIC-AVQA}   & Vision   & 0.333      & 0.823 \textcolor{ForestGreen}{\bf \scriptsize (+0.490)}        & \textbf{0.893} \textcolor{ForestGreen}{\bf \scriptsize (+0.560)}         \\
   & Audio    & 0.422      & 0.752 \textcolor{ForestGreen}{\bf \scriptsize (+0.330)}         & \textbf{0.821} \textcolor{ForestGreen}{\bf \scriptsize (+0.399)}        \\
\bottomrule
\end{tabular}
\end{table}

\FloatBarrier

Table~\ref{tab:slot_ablation} reports the companion surrogate-input
control used in Section~\ref{sec:proxy_contribution}. Zero and Gaussian features
reduce placeholder engagement, while valid modality features remain
closer to the real-feature trajectory. The cosines in this table are
trajectory similarities against the unmodified real-input run on the same
example, whereas Table~\ref{tab:logit_lens} reports similarity between
nearest-matched proxy--real pairs. The two quantities are measured
differently and their magnitudes are not comparable.

\begin{table}[!ht]
\centering
\caption{{Valid features engage the placeholder.}
We compare real features with surrogate inputs using the last-layer
residual cosine of the whole hidden-state trajectory against the
unmodified real-input run on the same example ($n{=}200$ pairs). This is
a different quantity from the nearest-pair cosine in
Table~\ref{tab:logit_lens} and the two should not be compared directly. Higher is closer to the real
trajectory. Zero and Gaussian inputs reduce engagement; valid features
remain closer.}
\label{tab:slot_ablation}
\small
\setlength{\tabcolsep}{7pt}
\renewcommand{\arraystretch}{1.10}
\begin{tabular}{@{}l c c c c c@{}}
\toprule
Modality      & bridged (ours) & zero          & Gaussian       & wrong-mod      & shuffled-real \\
\midrule
audio         & 0.291          & \textbf{0.097} & \textbf{0.136} & 0.309          & 0.316          \\
vision        & 0.321          & \textbf{0.208} & \textbf{0.237} & 0.391          & 0.432          \\
\bottomrule
\end{tabular}\\[2pt]
\end{table}

\section{Encoder selection}
\label{app:encoder_selection}
Our initial hypothesis when selecting encoders for our final model was that whichever encoders had the highest cosine similarity between text proxies passing through their text encoder and their modality counterparts would be the most useful for MSP. We calculated the mean cosine similarity between all COCO images and their text captions (Table~\ref{tab:encoder_cosine}).
\begin{table}[!ht]
\centering
\small
\caption{Mean cosine similarity between all COCO images and their text captions, used to inform encoder selection.}
\label{tab:encoder_cosine}
\begin{tabular}{@{}lr@{}}
\toprule
Vision encoder & Mean cosine similarity \\
\midrule
MetaCLIP B/16   & 0.407 \\
OpenAI CLIP B/16 & 0.347 \\
DFN2B B/16      & 0.331 \\
SigLIP2 B/16    & 0.152 \\
SigLIP1 B/16    & 0.143 \\
\bottomrule
\end{tabular}
\end{table}

Cosine similarity is not the only determinant of success with a given contrastive encoder, but we believe that it does matter for effectiveness of MSP training. One note is that sigmoid-loss models produce lower raw cosine values by design (SigLIP). For our final model that we report in the paper we utilized MetaCLIP due to its strong text-vision alignment and highest starting cosine similarity. This process was also repeated with a selection of audio encoders to inform our final model architecture.

\section{Anchor-supporting proxy examples}
\label{app:qualitative_appendix}

\begin{figure}[htbp]
  \centering
  \footnotesize
  \setlength{\tabcolsep}{4pt}
  \renewcommand{\arraystretch}{1.25}
  \begin{tabular}{>{\centering\arraybackslash}m{3.0cm}
                  >{\centering\arraybackslash}m{2.4cm}
                  >{\centering\arraybackslash}m{4.3cm}
                  >{\centering\arraybackslash}m{2.4cm}}
    \toprule
    \multicolumn{4}{@{}l}{\textbf{(a) MUSIC-AVQA} ~--~ \emph{audio anchor $\to$ vision proxy}} \\
    \midrule
    \textbf{Image} &
    \textbf{Real anchor (audio)} &
    \textbf{Generated proxy (vision)} &
    \textbf{Held-out GT} \\
    \midrule
    \includegraphics[width=2.8cm]{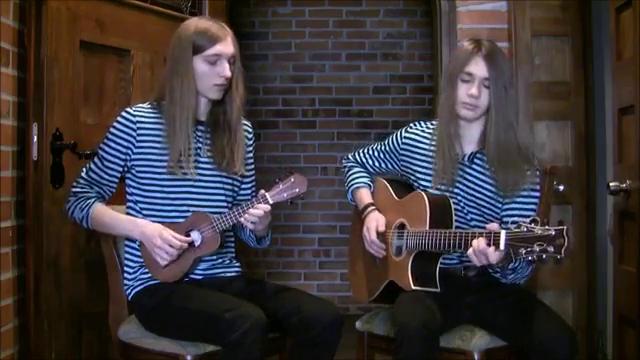} &
    \emph{Audio clip of a ukulele and guitar playing simultaneously.}\newline\vspace{2pt} &
    ``A red acoustic guitar and blue ukulele on a wooden stage. The guitarist strums chords while the ukulele player plucks strings.'' &
    Two musicians on stage; acoustic guitar and ukulele playing together. \\
    \midrule
    \multicolumn{4}{@{}l}{\textbf{(b) AVQA} ~--~ \emph{vision anchor $\to$ audio proxy}} \\
    \midrule
    \textbf{Image} &
    \textbf{Real anchor (vision)} &
    \textbf{Generated proxy (audio)} &
    \textbf{Held-out GT} \\
    \midrule
    \includegraphics[width=2.8cm]{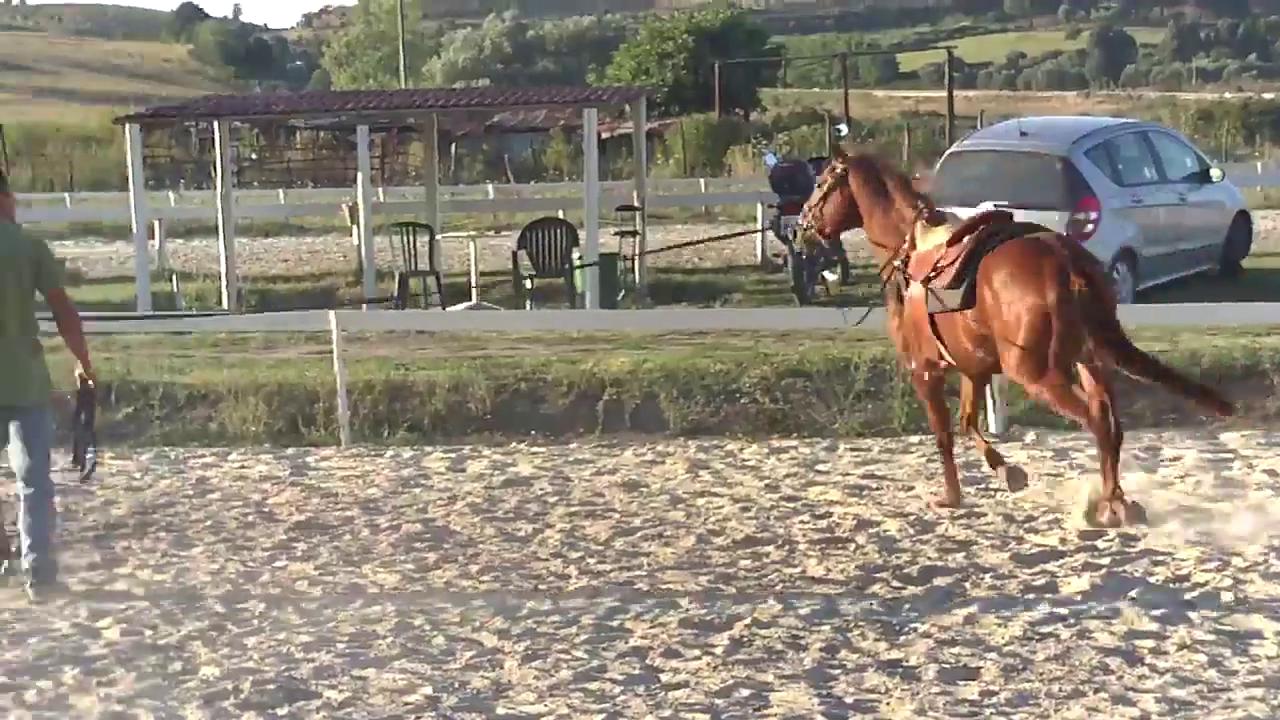} &
    \emph{Image of a brown horse in a sandy ring with a person leading it.} &
    ``A horse is neighing and hooves are clopping on the ground.'' &
    Horse neighing and hooves clopping on hard ground (audio held out). \\
    \bottomrule
  \end{tabular}
 \caption{{Examples of anchor-supporting proxy generation.}
  Each panel shows a real anchor in one modality and the proxy text
  generated for the other modality. The held-out
  ground truth is shown only for context; it is not used to generate
  the proxy.}
  \label{fig:qualitative_typeA}
\end{figure}

\FloatBarrier

\begin{figure}[!htbp]
  \centering
  \small
  \setlength{\tabcolsep}{8pt}
  \renewcommand{\arraystretch}{1.30}
  \begin{tabular}{@{}>{\centering\arraybackslash}m{4.2cm} m{8.0cm}@{}}
    \toprule
    \multicolumn{2}{@{}c@{}}{\textbf{(a) AVQA, sample \texttt{NumXu4eOqAw\_000006} --- white double-decker bus}} \\
    \midrule
    \includegraphics[width=4.0cm]{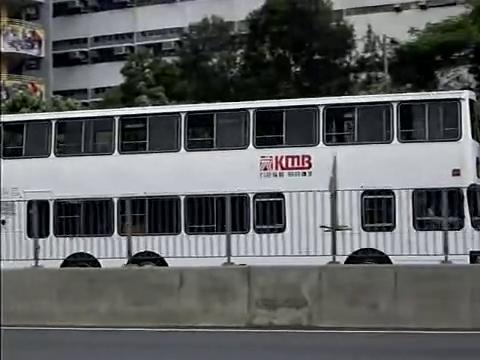} &
    \emph{Real audio anchor:}\, engine humming with tires rolling over pavement. \newline\vspace{3pt}
    \emph{Generated vision proxy:}\, ``A silver car speeds down a curvy mountain road at sunrise.'' \newline\vspace{3pt}
    \emph{Mismatch:} the real scene is a white double-decker bus on a highway; the proxy invents a silver car on a mountain road. Both descriptions are consistent with the engine $+$ tires audio. \\
    \midrule
    \multicolumn{2}{@{}c@{}}{\textbf{(b) AVQA, sample \texttt{xeAxybX3i1E\_000030} --- waterfowl}} \\
    \midrule
    \includegraphics[width=4.0cm]{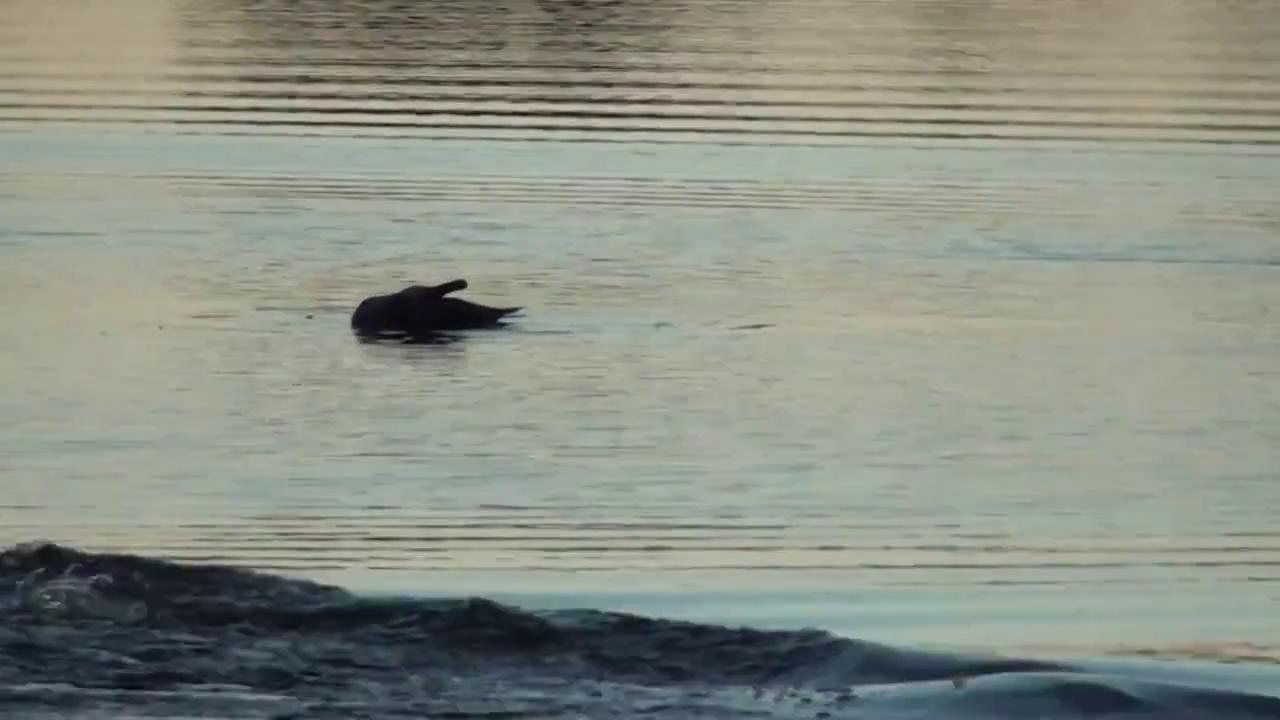} &
    \emph{Real audio anchor:}\, ducks calling near water. \newline\vspace{3pt}
    \emph{Generated vision proxy:}\, ``A dog grasping a duck.'' \newline\vspace{3pt}
    \emph{Mismatch:} the real scene shows ducks in water; the proxy describes a different scene that shares the duck object but introduces a dog. The audio (duck calls) remains consistent with both interpretations. \\
    \bottomrule
  \end{tabular}
  \caption{{Anchor-supporting proxy examples.} Generated vision
  proxies can differ from the held-out scene while remaining
  compatible with the real audio anchor.}
  \label{fig:qualitative_anchor_coherent}
\end{figure}

The qualitative figure (Figure~\ref{fig:qualitative_typeA})
shows faithful proxy generations, where the Phase-1 model correctly
describes the held-out modality. The two examples below (Figure~\ref{fig:qualitative_anchor_coherent}) illustrate a
more nuanced case: the proxy is not literally faithful to the
held-out scene, but remains consistent with the real anchor. MSP can
still benefit from these examples because the training signal is
anchor consistency rather than literal reconstruction of the held-out
modality, matching the shuffled-input result in Section~\ref{sec:proxy_contribution}.

\FloatBarrier

\clearpage
\section{Bridge variant comparison}
\label{app:bridge_variants}

Two calibration-time choices in Section \ref{sec:method} are validated by
ablation. The linear ridge bridge family beats identity, mean-shift,
and residual-MLP variants on held-out residual on both audio and
vision. Earlier co-trained bridge
variants collapsed toward the identity under the QA loss, so the
linear ridge is fit offline and frozen. Encoder-native terse prompt
templates beat literary 4-sentence templates by $+0.39$ post-bridge
cosine on audio (CLAP) and $+0.28$ on vision (MetaCLIP), so we adopt
the terse templates throughout Phase~2.

The deployed bridge of Section~\ref{sec:bridge-calibration} is a
ridge-regularized linear map (Eq.~\ref{eq:ridge_bridge}). For
completeness, Table~\ref{tab:bridges} compares four bridge families
on held-out residual $\varepsilon_{m'}=\mathbb{E}\|B_{m'}E_{m'}^{\text{text}}(c)-
E_{m'}^{\text{modality}}(o)\|_2$. The linear ridge family achieves the
smallest residual on both audio and vision.

\begin{table}[!htbp]
\centering
\caption{\textbf{Bridge calibration residuals.} Held-out bridge error
from text-tower to modality-encoder space; lower is better. The
linear bridge gives the lowest residual for audio and vision.}
\label{tab:bridges}
\small
\setlength{\tabcolsep}{8pt}
\renewcommand{\arraystretch}{1.10}
\begin{tabular}{@{}l c c@{}}
\toprule
Bridge                                   & Audio        & Vision       \\
\midrule
identity (no bridge)                     & 0.94         & 8.22         \\
mean-shift                               & 0.86         & 5.36         \\
residual MLP                             & 0.78         & 2.12         \\
\textbf{linear} (ridge) $\checkmark$ & \textbf{0.70} & \textbf{0.86} \\
\bottomrule
\end{tabular}\\[2pt]
\end{table}

\FloatBarrier

\section{Additional composition controls}
\label{app:rebuttal_controls}
All numbers below are from the reported Qwen3-8B LoRA model at $r{=}512$, on the test splits of the given benchmarks.

\paragraph{Training baselines.}
In Table~\ref{tab:training_methods_comparison}, the ``without proxy and bridge'' row omits the non-anchor modality entirely, meaning that no token is given, and it is not zeroed or replaced by a learned null. The ``without bridge'' row keeps the same proxies, but injects them without the learned text-modality mapping the bridge contributes.

\begin{table}[htbp]
\centering
\small
\caption{QA supervision and MSP composition. All numbers are at $r{=}512$ on the test splits.}
\label{tab:training_stages}
\begin{tabular}{@{}lrr@{}}
\toprule
Stage & AVQA & MUSIC-AVQA \\
\midrule
Phase 1 only (per-modality projectors) & 70.21 & 45.75 \\
+ QA supervision only, no proxies, no bridge & 77.46 & 55.18 \\
+ full MSP composition & 85.58 & 65.65 \\
\midrule
QA-supervision contribution & +7.25 & +9.43 \\
composition contribution & +8.12 & +10.47 \\
\bottomrule
\end{tabular}
\end{table}

\FloatBarrier

\paragraph{Real modalities vs. proxy modalities at inference.}
We observe that for both benchmarks, highest accuracy is obtained when both modality inputs are real modality samples. Additionally, we see consistent decreases in performance when a single modality is swapped for a proxy, and an even larger decrease when both modality inputs are swapped for generated proxies.
All six drops are significant (all $p<10^{-6}$, paired McNemar test over per-sample predictions from a fixed model).
These results are consistent with our view that proxies function as a useful training-time scaffold for learning cross-modal structure, but do not carry the same fidelity of information as their real counterparts.

\paragraph{Token-matched controls.}
The Gaussian is matched to the real features' per-dimension mean and standard deviation, and shuffled is a genuine encoder output (just one taken from a different test example). Notably, real examples with the wrong features cost the most on both controls, which suggests that a real sample from the same distribution can actively mislead a model when its content does not help answer the given query.

\section{MUSIC-AVQA-R}
\label{app:musicavqar}
As an additional control, we tested our model against MUSIC-AVQA-R~\citep{ma2024musicavqar}, which is the debiased re-split of MUSIC-AVQA that contains a tail reweighted toward rarer answers, with the specific goal of defeating answer-prior exploitation. On the full $211{,}572$-question test set we achieved the following (Table~\ref{tab:musicavqar}):

\begin{table}[htbp]
\centering
\small
\caption{MUSIC-AVQA-R results on the full $211{,}572$-question test set.}
\label{tab:musicavqar}
\begin{tabular}{@{}lrrr@{}}
\toprule
Split & $n$ & MSP & Majority class \\
\midrule
head & 138,153 & 67.26 & 35.60 \\
tail & 73,419 & 47.76 & 31.67 \\
overall & 211,572 & 60.49 & --- \\
\bottomrule
\end{tabular}
\end{table}

We believe that this result shows additional evidence that MSP learns to use cross-modal structure effectively and can perform well even when compared against models that use direct supervision.

\section{Sensitivity to proxy quality at test time}
\label{app:proxy_quality}
We compare the model at its best (all real inputs) with proxies at two levels of quality: (1) scene-grounded proxies are proxies generated from the Phase 1 model by having it observe an anchor modality and imagine what the counterpart modality might be, and (2) a question-derived proxy with no accompanying modality input. A question-derived proxy is generated by asking the model for a possible audio proxy based on the given question's prompt (Table~\ref{tab:proxy_quality}).

\begin{table}[htbp]
\centering
\small
\caption{Audio proxy quality at test time with real vision.}
\label{tab:proxy_quality}
\begin{tabular}{@{}lr@{}}
\toprule
Audio proxy quality (vision real) & AVQA ($r{=}512$) \\
\midrule
real audio & 85.58 \\
scene-grounded (imagined from real vision) & 81.19 \\
question-derived (observes nothing) & 76.15 \\
\bottomrule
\end{tabular}
\end{table}

For example:
\begin{quote}
\textbf{Question:} ``What's driving in the video?''\\
\textbf{Audio:} \texttt{<proxy>} A car engine roars as tires screech and glass shatters \texttt{</proxy>}\\
\textbf{Vision:} \texttt{<real>} A blue car drives past a red truck on a sunny highway \texttt{</real>}
\end{quote}

From the token-matched controls, the table above, and this example (generated by the 8B $r{=}512$ model) it is clear why weaker proxies might not function as well as stronger counterparts. In this example there is clearly conflicting information---the audio paints the picture of a car racing and creating some kind of accident, while the vision input specifies a more peaceful highway scene.

This also implies that a multimodal model may not always be necessary for generating high-quality proxies if the generator has a decent enough signal (e.g., detailed captions as input) and sufficient multimodal priors learned from text. This is a future direction worth exploring.

\clearpage
\section{Paired data at a smaller scale}
\label{app:paired_pilot}
For this control we took a text-only Qwen3 4B model and finetuned it on AVQA across 5 training paradigms: (1) paired data only, (2) 1 anchor + 1 proxy (general MSP recipe, alternating the anchor modality), (3) interleaved unimodal (1 modality per sample, other slot is left empty), (4) vision only data, and (5) audio only data (Table~\ref{tab:paired_pilot}).

\begin{table}[!ht]
\centering
\small
\caption{Paired data at a smaller scale: a text-only Qwen3 4B model finetuned on AVQA.}
\label{tab:paired_pilot}
\begin{tabular}{@{}lr@{}}
\toprule
Method & AVQA accuracy \\
\midrule
Paired upper bound (fully paired data) & 87.45 \\
MSP (ours) & 86.00 \\
Interleaved unimodal (alternating modalities by sample) & 81.46 \\
Vision only & 80.72 \\
Audio only & 75.35 \\
\bottomrule
\end{tabular}
\end{table}

While this does not determine the upper bound of our headline $r{=}512$ model, it does indicate MSP is a strong alternative to fully paired data. The important takeaway is not just the gap between fully paired and MSP (which is relatively small), but the gap between MSP and interleaved unimodal training. This shows that when fully paired data is unavailable, there is an option beyond leaving absent modality slots empty.

\ifarxiv\else
\clearpage
\newpage
\input{checklist.tex}
\fi

\end{document}